\documentclass[article,preprint,pdftex,moreauthors]{Definitions/mdpi}

\firstpage{1} 
\pubvolume{1}
\issuenum{1}
\articlenumber{0}
\pubyear{2025}
\copyrightyear{2025}
\datereceived{ } 
\daterevised{ } 
\dateaccepted{ } 
\datepublished{ } 
\datecorrected{} 
\dateretracted{} 
\hreflink{} 
\pdfoutput=1 

\Title{A Visual Tool for Interactive Model Explanation using Sensitivity Analysis \small This work is a preprint version of a paper currently in preparation with co-authors.}

\TitleCitation{A Visual Tool for Interactive Model Explanation using Sensitivity Analysis}

\Author{Manuela Schuler $^{1,2,*}$\orcidA{}}
\AuthorNames{Manuela Schuler}
\AuthorCitation{Schuler, M.}

\address{%
$^{1}$ \quad German Research Center for Artificial Intelligence (DFKI), Saarbrücken, Germany\\
$^{2}$ \quad Saarland University, Saarbrücken, Germany
}

\corres{Correspondence: Manuela.Schuler@dfki.de}

\abstract{We present SAInT, a Python-based tool for visually exploring and understanding the behavior of Machine Learning (ML) models through integrated local and global sensitivity analysis. Our system supports Human-in-the-Loop (HITL) workflows by enabling users—both AI researchers and domain experts—to configure, train, evaluate, and explain models through an interactive graphical interface without programming. The tool automates model training and selection, provides global feature attribution using variance-based sensitivity analysis, and offers per-instance explanation via LIME and SHAP. We demonstrate the system on a classification task predicting survival on the Titanic dataset and show how sensitivity information can guide feature selection and data refinement.}


\begin{document}

\begin{figure}[ht]
  \includegraphics[width=\textwidth]{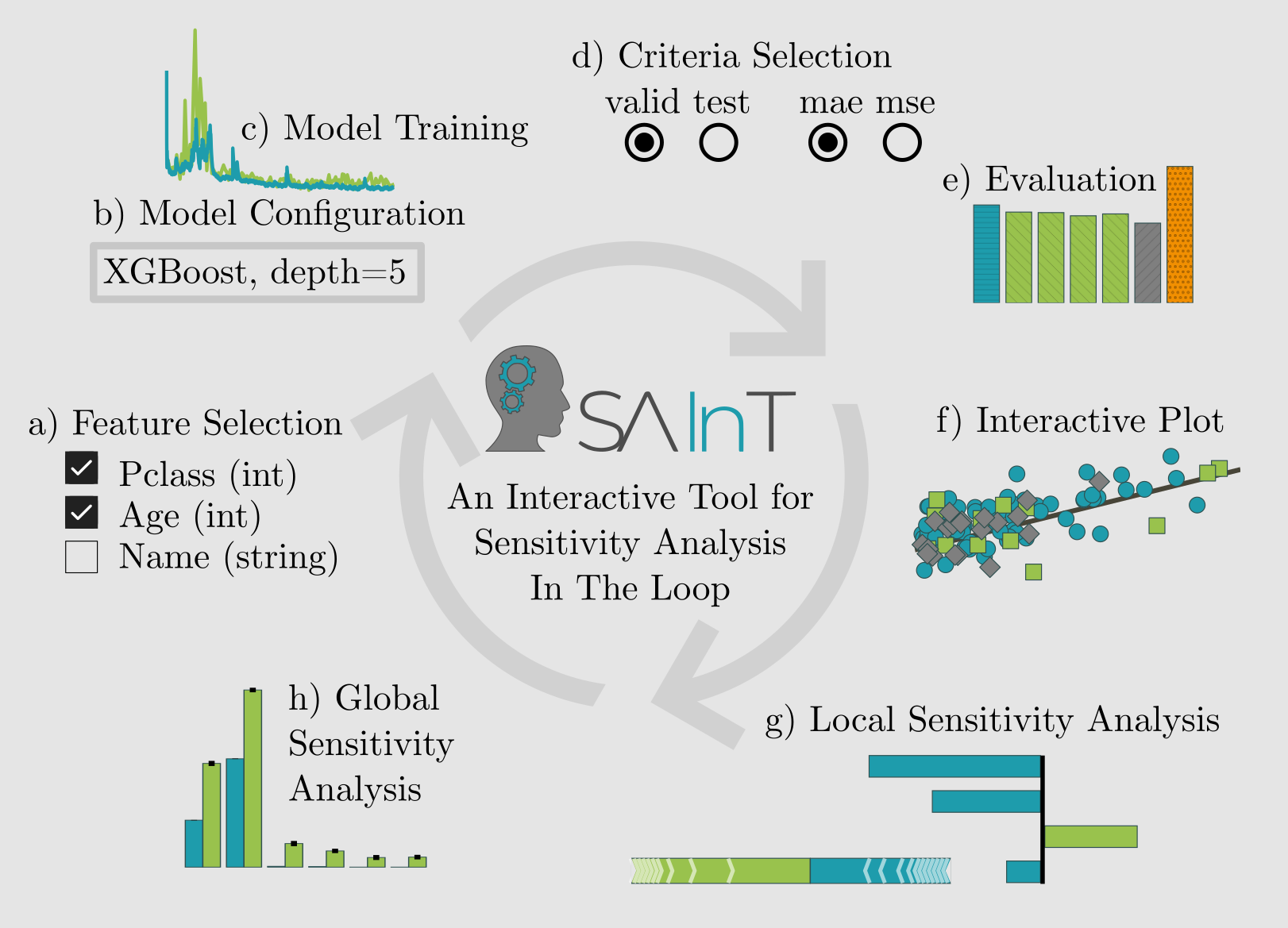}
  \caption{SAInT integrates Interactive Machine Learning (a-e) and Sensitivity Analysis (f-h) in a loop with user interactions. a) Feature Selection: Define input and output features. b) Model Configuration: Configure model parameters. c) Model Training: Train or load models. d) Criteria Selection: Select dataset and loss. e) Evaluation: Error plot of all models. The best model is selected. f) Interactive Plot of the best model: Click onto a sample. g) Local Sensitivity Analysis: Local explanations for the selected sample. h) Global Sensitivity Analysis: Identify the best features, which can be used in a) as refinement.
  \label{fig:teaser}}
\end{figure}

\section{Introduction}
Understanding how Machine Learning (ML) models reach decisions is crucial to foster trust and promote meaningful human-machine collaboration.
Local Sensitivity Analysis (LSA) methods provide situation-specific interpretability, revealing the details of individual predictions.
Global Sensitivity Analysis (GSA) provides a wider perspective on feature importance across the entire dataset, considering feature interactions.
This paper presents SAInT, a novel Python tool that integrates these techniques into a graphical user interface, supporting training, inference, and explanations of ML models in an end-to-end manner.

In interdisciplinary AI projects, effective communication between AI researchers and domain experts is crucial.
AI researchers often present complex model outcomes to domain experts who have the required understanding of the domain and data, but lack programming expertise.
This disparity in professional backgrounds  can lead to misunderstandings.
Our proposed tool, SAInT, addresses this problem. The tool empowers domain experts to inspect problematic cases, evaluate the performance of models on other datasets, draw conclusions about model reliability, and improve the data generation for retraining improved models.
AI researchers can use the tool to tune hyperparameters, detect data biases of sensitive features, and to gain insights in the model decision process and the influence of features on the prediction.
 
\subsection{Visual Analytics}
Visual Analytics (VA) is the idea to use visualization software to understand data and support decision making.
Recently, several new methods were suggested for handling large-scale and high-dimensional datasets \cite{Moritz2024}, and for leveraging advancements in ML and interactive visualization \cite{Heer2024}.
The integration of augmented reality and immersive technologies \cite{Hammady2020}, \cite{Zheng2017}, \cite{Chandra2019} has opened new possibilities for data exploration, providing users with an enhanced and interactive experience.
Notable progress was made on VA approaches that address the challenges associated with streaming data and real-time analytics \cite{Fischer2011}, \cite{Chung2016}, \cite{Eaglin2017}.

Additionally, the integration of VA with Explainable Artificial Intelligence (XAI) has gained importance for enhancing the interpretability of complex ML models \cite{Alicioglu2022}, \cite{Hohman2018}.
One focus is explaining neuron layers of Convolutional Neural Networks \cite{Liu2017}, \cite{Schorr2021}.
Current research in VA with XAI is largely focused on visual explanations for time series \cite{Schlegel2023_A_Deep_Dive}, \cite{Schlegel2023_Visual_Explanations} and explanation visualizations for Large Language Models (LLMs) \cite{Brath2023}.

\subsection{Sensitivity Analysis}
Sensitivity Analysis (SA) is a XAI technique to attribute the uncertainty in the output of a function or model to different sources of input uncertainty and determine the most important influence factor \cite{Saltelli2002}.
SA approaches can be divided into two categories: local (LSA) and global (GSA).

LSA can be understood as the partial derivative of a model output with respect to a model input \cite{Wei2013}.
Local Interpretable Model-Agnostic Explanations (LIME) \cite{Ribeiro2016} and SHapley Additive exPlanations (SHAP) \cite{Lundberg2017} are local approaches focusing on explaining predictions of individual instances and offering insights into the contribution of features for specific data points.
Despite being intuitive, LSA only provides a localized view of the problem space for a specific situation \cite{Razavi2021}.

GSA quantifies the effect of inputs to output uncertainty by fixing them over their full distribution ranges \cite{Wei2013}.
Variance-based approaches \cite{Morris1991}, \cite{Sobol2009}, \cite{Wei2013} average many local base points to achieve global sensitivity.

The Extended Fourier Amplitude Sensitivity Test (eFAST) \cite{Saltelli2012} is a computationally efficient variance-based method that approximates the first and total order Sobol indices by decomposing the variance of the model output using Fourier analysis.
Distribution-based approaches use either higher-order moments than variance, e.g. skewness or kurtosis \cite{DellOca2017}, or are moment-independent by measuring the difference between the unconditional distribution of the output and its conditional counterparts when one or more inputs are fixed \cite{Razavi2021}, \cite{Borgonovo2007}, \cite{Pianosi2018}.

Variogram-based approaches generalize global methods by introducing variograms and covariograms \cite{Razavi2016}, \cite{Sheikholeslami2020}, \cite{Becker2020}.
The VARS method \cite{Razavi2019} proposed as a computationally efficient replacement for Sobol indices was found to be a complementing, rather than replacing method by Puy at el. \cite{Puy2021}, due to its lack of a clear definition of feature importance.

Regression-based methods \cite{Friedman1991} depend on prior assumptions about the model response form and may be unreliable if the fit is poor.
Response-surface methods use Bayesian estimation \cite{Pronzato2019} or Polynomial Chaos Expansion (PCE) and Gaussian Process Regression \cite{Wang2020}.

There are several tools for performing GSA: SALib \cite{Herman2017}, \cite{Iwanaga2022} is a well-documented Python tool with various GSA methods, but lacks performance optimization for large datasets.
The R tool sensobol \cite{Puy2022} focuses on Sobol indices and is efficient for large-scale models.
SAGE Tool \cite{Covert2020} specializes in ML models.

SAFE \cite{Noacco2019} is a user-friendly Tool similar to SALib.
VISCOUS \cite{Sheikholeslami2021} emphasizes visualization, but is complex and has a steep learning curve.
VARS-TOOL \cite{Razavi2019} provides detailed insights into model behavior, but also has a steep learning curve.

These tools help to analyze the sensitivity of model outputs to variations in input parameters across the dataset and provide a full understanding of the impact of input parameters on model predictions.

\subsection{Human-in-the-Loop}
Human-in-the-Loop (HITL) approaches \cite{Wang2022}, \cite{Holzinger2021} integrate human expertise into ML processes, involving feedback, validation, and decision-making.
The ``Diff in the Loop`` concept \cite{Wang2022} visualizes changes in data resulting from changes in code in exploratory data analysis. Multi-modal causability can be achieved using Graph Neural Networks \cite{Holzinger2021}.
HITL technologies can be part of the pipeline involving data processing, model training and inference and system construction and application \cite{Wu2021}.
Some HITL methods integrate human knowledge into modeling \cite{Kumar2019}, \cite{Hartmann2018}, \cite{Chen2019}, \cite{Lin2020}: Kumar et al. \cite{Kumar2019} explore user control in HITL topic models, demonstrating how human feedback can refine model outcomes. Hartmann et al. \cite{Hartmann2018} introduce deep reinforcement learning combined with prior human knowledge for optimizing velocity control in autonomous systems. Chen et al. \cite{Chen2019} propose a deep learning method that integrates prior knowledge and temporal data to enhance decision-making in autonomous driving. Lin et al. \cite{Lin2020} present a novel deep learning framework that incorporates Hough-Transform line priors to improve line detection in images.
According to a recent literature survey \cite{Wu2021}, most HITL research focuses on natural language processing.
One reason is that it is harder for users to directly interact with images (except for labeling) \cite{Wu2021}.
Multi-modal systems are realized with inverse reinforcement learning \cite{Arora2018} or with Graph Neural Networks \cite{Holzinger2021} for information fusion.
Obtaining essential samples and labeling them with human intervention is critical in HITL technology \cite{Wu2021}.

\subsection{Interactive Machine Learning}
Interactive Machine Learning (IML) is a research direction within HITL and comprises algorithms and user interface frameworks that facilitate ML with human interaction. The field has a long history with early works emphasizing the collaboration between humans and machines in the learning process \cite{Fails2003}, \cite{Knox2009}.
Crayons \cite{Fails2003} is one of the first tools that uses an IML process for creating image classifiers with human guidance. Knox et al. \cite{Knox2009} propose the TAMER framework, which enables agents to learn behavior from real-time human reinforcement rather than algorithmic feedback.
Recently, interest in understanding model behavior has increased.
ScatterShot \cite{Wu2023} is an interactive tool for improving the fine-tuning of Large Language Models (LLMs) using a specific sampling technique.

Our work is closest related to the What-If Tool \cite{Wexler2019} and ExplAIner \cite{Spinner2020}:
The What-If Tool is an IML tool that lets users probe and understand the behavior of ML models on classification and regression tasks and can be integrated into TensorBoard or Jupyter Notebook.
It allows for custom prediction functions.
It is flexible but integration requires  programming skills.
ExplAIner is another IML tool that explains model predictions and can also be integrated in TensorBoard.
It lets users explore and understand the decisions made by ML models.
Both tools take a model-centric view and allow the analysis of concrete ML systems.

\subsection{Our Contribution}
We introduce SAInT, our Interactive Tool for Sensitivity Analysis in The Loop. The tool abstracts the ML system by the combination of three steps: (1) automatic training of a collection of ML models, (2) automatic selection of the best model, and (3) sensitivity analysis of the selected model. From a user perspective, this leads to a data-centric approach and allows to make statements on the properties of datasets, rather than models. As a consequence, AI becomes accessible to domain experts as a tool for interactive data analysis.

\section{Materials and Methods}
\subsection{Overview}
The SAInT tool implements a Human-in-the-Loop workflow for data understanding. A user configures, trains and evaluates several machine learning models on a dataset, then performs local and global sensitivity analyses to understand feature importance (Figure~\ref{fig:teaser}). The tool 
consists of the following components: Feature selection including data loading, model training or loading, model evaluation and automated model selection, an interactive visualization with integrated LSA for in-depth model explanation and GSA for identification of important input features.
Our tool handles preprocessing of CSV-data and supports classical models via scikit-learn \cite{Scikitlearn2011} and deep learning models via fastai \cite{Fastai2020}.
The graphical user interface is implemented using the Dash Plotly Python library \cite{Plotly2015}.
GSA Sobol indices on the dataset are computed using the eFAST implementation of SALib \cite{Herman2017}, \cite{Iwanaga2022}.
LIME \cite{Ribeiro2016} and SHAP \cite{Lundberg2017} are integrated as LSA components and can be computed per data sample.
SAInT supports single- and multi-output regression and classification, both linear and non-linear, for tabular CSV data with int, float, or string types. String data is converted to binary features via one-hot encoding.

\subsection{Feature Selection and Data Loading}
The data is provided in CSV file format.
Users can either provide one single CSV dataset with all data or the data as separate files.
In the first case, the entire dataset can be randomly divided into train, validation, and test data with customizable percentage values.
If the data should not be split randomly, we recommend arranging the data beforehand and providing separate CSV files. The user selects regression or classification mode, optional normalization (min-max or mean-std), and specifies features as either input or output.
We provide optional data balance analysis and support for applying balancing to specific inputs and outputs.

\subsection{Model Loading and Training}
Users can load or train new models at any time.
SAInT supports classical ML models such as RandomForest and XGBoost \cite{Chen2016} and Deep Learning models like Multilayer-Perceptron (MLP) or Tabular ResNets.
Models are trained per hyperparameter combination and saved to disk.

\subsection{Model Evaluation and Automated Model Selection}
Following loading or training a model, all loaded models are evaluated on the train, validation, or test dataset and the error plot for all models is shown.
Users can choose among different loss functions for regression and classification \cite{Ciampiconi2023}.
Loss functions for regression: Mean Absolute Error (MAE), Mean Squared Error (MSE), Root Mean Squared Error (RMSE), Mean Squared Logarithmic Error (MSLE), Root Mean Squared Logarithmic Error (RMSLE), Log-cosh Loss.

For classification tasks, the following loss functions are supported: Hinge Loss, Smoothed Hinge Loss, Squared Hinge Loss, Modified Huber Loss, Ramp Loss, Cross Entropy Loss, Binary Cross Entropy Loss, Negative Log-Likelihood Loss.

After each evaluation, the model with the lowest error on the chosen dataset is selected for further analysis.
By switching between different datasets, users can explore the generalization ability of the model and verify if the best model on the validation dataset also performs well on the test dataset.
SAInT can be used for hyperparameter tuning:
Users can gradually select the best model type and add new models with adapted hyperparameters to search for better models for the current task.

A typical approach is to first train a default XGBoost and Random Forest model and compare them in terms of performance. Then, a suitable maximum depth could be found by iteratively training new models of the better-performing model type with an increased or decreased depth and comparing again.

Next, neural networks are added for comparison. Starting with a simple architecture of two hidden layers with 64 and 32 neurons, experiments should incrementally modify neuron sizes, dropout rate, and training epochs. The tabular ResNet has the parameters layer size and the number of residual blocks which should start at 2 residual blocks and a layer size of 64. Parameters should be increased incrementally until the best setup for the task is found.

Direct model comparison is limited to those trained on the same set of input and output features, as feature selection is part of the GUI pipeline.

\subsection{Identification of Important Input Features}
A GSA method can be used to calculate the importance of features for the selected model on the entire data space.
This starts automatically after identifying the best model.
The GSA plot shows first and total order Sobol indices (y-axis) for each input feature (x-axis).
The first order Sobol measures the main effect of a single input parameter on the output, ignoring interactions with other parameters.
The total Sobol measures the total effect of a single input parameter, including its interactions with other parameters.
A high difference between the first and total Sobol values indicates high interdependence with other features by increased feature interaction.

\subsection{Interactive Visualization and Local Explanations}
For each output feature, one interactive subplot is generated.
The default plot shows color coded ground truth and prediction values of each sample.
Data is dynamically visualized and can optionally be sorted based on ground truth values or according to prediction values.
Alternatively, users can switch to a Goodness-of-fit visualization, which is useful for regression mode:
The x-axis corresponds to predictions, while the y-axis corresponds to ground truth values.
The diagonal indicates the optimal fit, i.e. prediction equals ground truth.
This plot shows how well the model predictions align with the actual ground truth values and helps detect outliers with high deviation from the ground truth.

In the interactive plot, individual samples can be selected, which triggers the computation of local explanations with optionally LIME or SHAP.
The LIME explanation shows the prediction probabilities, negative and positive impact of features and the input feature values.
For numerical features, lower and upper decision thresholds are also given.

The SHAP visualization shows a force plot.
Features on the left side have a positive influence on the model prediction, features on the right have a negative influence.
Additionally to the optional LIME and SHAP explanations, the prediction, ground truth and features are shown in the popup window.
Users can use radio buttons to toggle between normalized and denormalized data.

\subsection{Sensitivity Analysis in The Loop}
Our tool can be used by different users, each having a different background and expertise.

They can use our tool to tune hyperparameters, detect data biases, gain insights in model decision-making, select features, visualize outliers, apply models on other datasets, explore model reliability, and improve data generation for model refinement.

Hyperparameter Tuning: When a model is added, it is evaluated and compared to other models. The best model for a dataset is identified by iteratively updating parameters and assessing error changes.

Bias Detection: The user can measure the impact of sensitive features, such as gender or race, on the model. Since the model replicates the biases in the training data, this helps in identifying and addressing biases.

Gaining insights in model decision-making: Using LSA on specific samples within the interactive graph, the user can determine which features are most important for predictions with high or low output values.

Features selection: Using GSA, the user can identify important features and remove unimportant input features in the next training iteration.

Visualizing outliers: The user can identify and visualize outliers, which can be analyzed to improve the generation of new training data.

Application to other datasets: The user can evaluate model files on own datasets.

Explore model reliability: The user can explore the model behavior in different data scenarios, building trust in the trained model.

Data generation: The user can improve the data generation process. By examining problematic samples, the user can detect areas, where the model has difficulties in prediction.
New data samples can be added to the training dataset and an improved model can be trained.

\section{Results}
In the following, a survival prediction on the Titanic dataset is provided as a classification use case \cite{Cukierski2012}.
A Kaggle dataset was chosen for its accessibility, recognition within the Machine Learning community, ease of understanding, and alignment with user expectations regarding results.
We also provide information about the runtime performance of our tool.

\begin{figure}[ht]
\includegraphics[width=\textwidth]{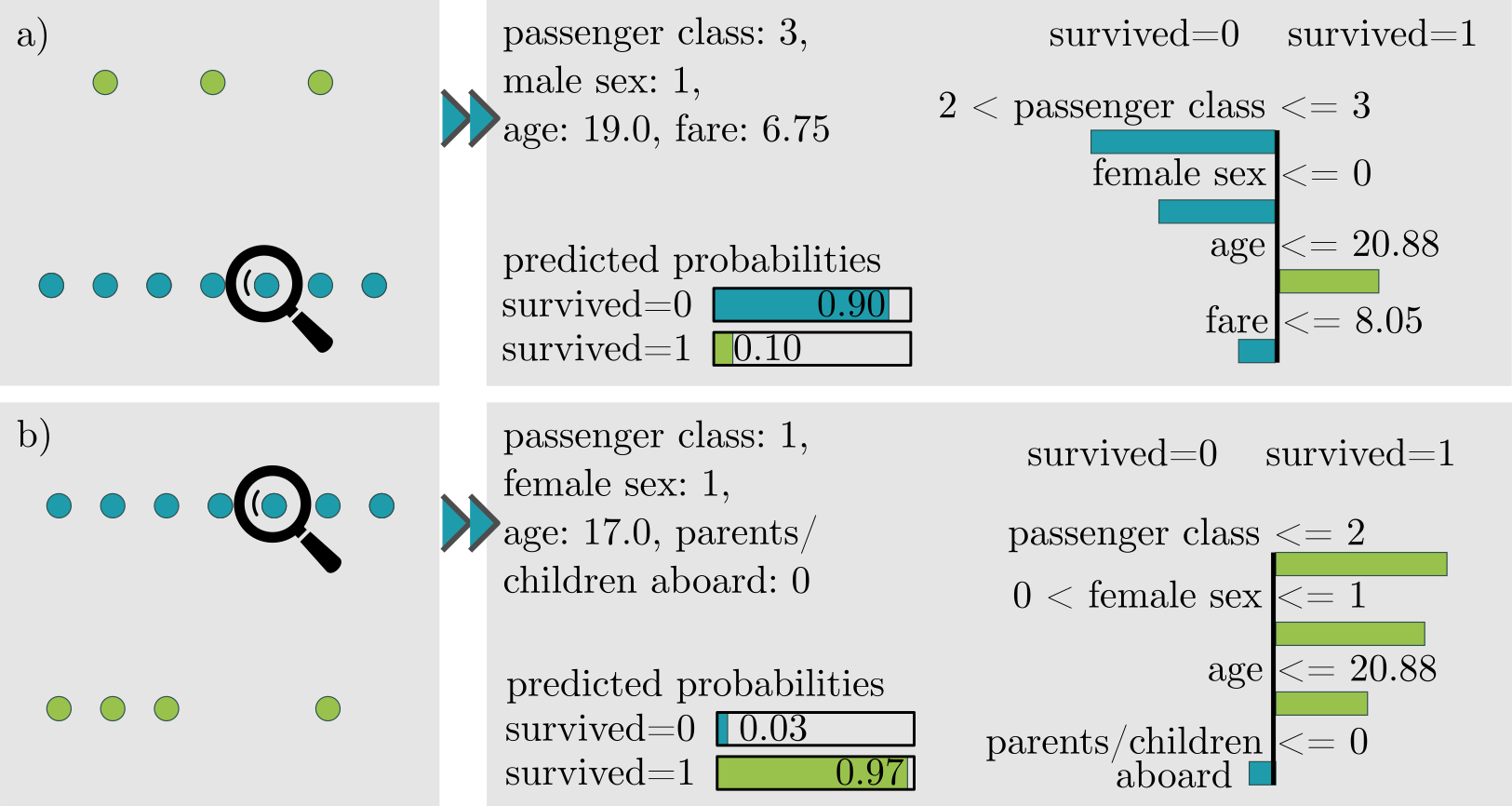}
\caption{LIME explanations for survival prediction (classification) on the Titanic dataset a) Titanic non-survivor: {\color{black}{Passenger}} class 3 and male sex are most influential. b) Titanic survivor: {\color{black}{Passenger}} class 1 and female sex are most influential.\label{fig:titanic_lsa_dataset}}
\end{figure}

\subsection{Use Case: Survival prediction on the Titanic Dataset}
We demonstrate how GSA can be used to reduce the number of features of a model. We train a model on the ``Titanic - Machine Learning from Disaster'' \cite{Cukierski2012} Kaggle dataset to predict if a passenger survived.
The file contains 891 passenger data samples.
The selected input features are: passenger class, sex, age, siblings or spouse aboard, parents or children aboard and fare. While the term ``gender'' is often preferred in general usage, we use the term ``sex'' here to align with the terminology in the dataset, which uses ``sex'' as a column label. This distinction reflects the dataset's design and does not imply a broader preference for terminology.
Preprocessing steps include one-hot-encoding for the categorical feature ``sex'' and excluding data samples containing missing data values in the selected input features.

\begin{figure}[ht]
\centering
\includegraphics[width=1.0\textwidth]{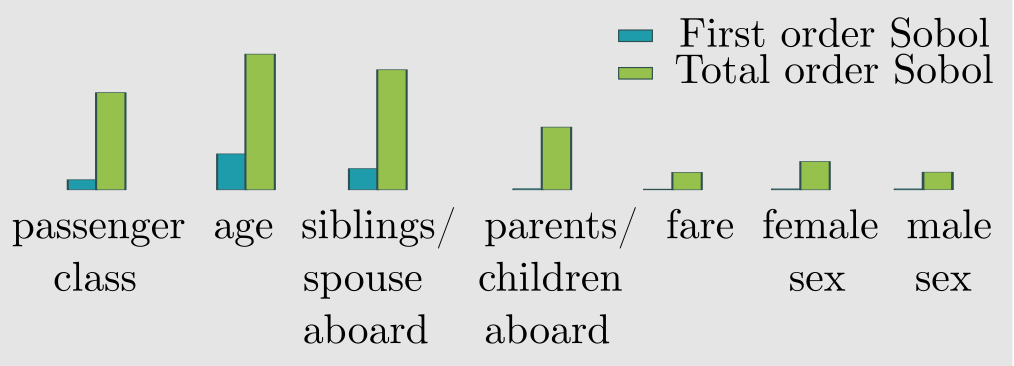}
\caption{Most important features identification with GSA: Survival prediction on the Titanic dataset: Passenger class, age, and siblings or spouse aboard are most important. Fare, parents or children aboard and sex (female sex, male sex) have less effect.\label{fig:gsa}}
\end{figure}

Figure~\ref{fig:titanic_lsa_dataset}a) and Figure~\ref{fig:titanic_lsa_dataset}b) show local explanations for two samples of the validation dataset:
A 19-year-old man in passenger class 3 did not survive, whereas a woman of age 17 in passenger class 1 did survive.
Female sex, traveling first class and an age up to 20 years is beneficial for the chance to survive, whereas male sex and all other passenger classes are decreasing factors.
The main local factors contributing to the survival chance are passenger class and sex.
According to the GSA in Figure~\ref{fig:gsa} the globally most influencing features are passenger class, age and siblings or spouse aboard.
Fare and parents or children aboard contribute the least.
The information of the GSA can help to reduce the number of input features:
We train a new model excluding the two features with lower GSA influence.
This leads to a redistribution of the GSA onto the remaining features passenger class, sex, age, siblings or spouse aboard, allowing the model to focus on the most influential features and better capture relevant predictive patterns.

\subsection{Performance Evaluation}

We verified that large data files can be processed efficiently on a low-end consumer laptop. Loading a CSV file with 100,000 samples took approximately 250 ms, training an XGBoost model on the dataset took 1.2 seconds, computing the model error for all data points took 450 ms, and performing GSA of the selected model took 900 ms.

\section{Discussion}
Most IML tools have a model-centric approach. Our tool abstracts the machine learning system into three automated steps, creating a data-centric approach: model training with any possible model configuration, automated model selection, followed by sensitivity analysis. This enables users to gain insights on the properties of datasets, making AI more accessible for domain experts as a tool for interactive data analysis.
\subsection{Strengths}
SAInT simplifies the process of constructing ML models, especially for users with limited programming experience. It can be used on regression and classification tasks.

As a practical tool for AI researchers and domain experts of other areas, SAInT visually explains model performance on selected samples using interactive plots.
Users can check instances where the model delivers good performance and investigate  inaccurate samples, offering valuable insights into the behavior of the model.
Our tool can be used to validate the performance of a model on alternative test datasets.
It can help to uncover biases that are hidden in the data.
SAInT integrates GSA and LSA visually, which is an intuitive way for interpretation of feature importance.

SAInT is a standalone application with a graphical user interface, such that all data is stored locally, preventing the data privacy concerns related to online tools.

\subsection{Limitations and Future Work}
Our tool is limited to tabular data as CSV files, but extending it to image and text data is straight forward in principle.
SAInT currently supports well known loss functions, but we plan to extend it by user-defined loss functions in future.
Users seeking to train models with different inputs must create a new user folder with identical data, as models with varying input features are not supported during evaluation.
Future iterations of the tool could incorporate dynamic support for adjusting input and output feature configurations during runtime.
Model selection is automatic based on the lowest error in the selected dataset.
In a future version, users will be able to select models by clicking on the error plot and compare different models.
Our tool is limited to single device training using FastAI.
If there is a strong linear correlation between input features, the response of the sensitivity analysis will be split between the correlating features, diluting the result. The system detects this situation and raises a warning. If feature correlation is undesirable, the features can be transformed to a lower-dimensional coordinate system without linear correlation using a Principal Component Analysis (PCA) as preprocessing step.
We demonstrated the potential of our software through a use case, highlighting its integration of methodologies such as LIME, SHAP, and eFAST, which provide interpretable insights into model behavior. While these results underscore the utility of our tool, future work will include a comprehensive user study to further validate its effectiveness and usability.
\section{Conclusions}
HITL methods are an intriguing approach to combine human expertise with ML processes.
The concept is beneficial to feedback, validation, and decision-making.
In our tool SAInT, we incorporate LSA and GSA, providing human expertise for nuanced model evaluation and parameter tuning in successive iterations.
It features ready-to-use analysis techniques including LIME and SHAP for LSA and eFAST for GSA.
We offer users direct feedback through our IML process.

SAInT enables users to detect biases in unfair model decisions, start a new training with reduced input features based on GSA, and detect the main reasons for a high or low survival chance.
We also demonstrated that the user could interact with hyperparameters with instant feedback on prediction performance of the model.

\vspace{6pt} 

\supplementary{
The Titanic dataset is available at \url{https://www.kaggle.com/competitions/titanic/data} (accessed on 06 August 2025). A video of how to use our tool is available at \url{https://www.youtube.com/watch?v=m269sWdYUUI} (accessed on 06 August 2025).}

\funding{This research was funded by the German Federal Ministry of Education and Research (BMBF) grant number FKZ 01IS21106 (ENGAGE).}

\acknowledgments{We would also like to thank K-UTEC AG SALT TECHNOLOGIES for successful collaboration, since the idea for the usefulness of such a tool came up during a joined industry project.}

\newpage

\abbreviations{Abbreviations}{
The following abbreviations are used in this manuscript:\\

\noindent 
\begin{tabular}{@{}ll}
AI & Artificial Intelligence\\
ML & Machine Learning\\
HITL & Human-in-the-Loop\\
VA & Visual Analytics\\
IML & Interactive Machine Learning\\
XAI & Explainable Artificial Intelligence\\
SA & Sensitivity Analysis\\
LSA & Local Sensitivity Analysis\\
GSA & Global Sensitivity Analysis\\
LLM & Large Language Model\\
LIME & Local Interpretable Model-Agnostic Explanations\\
SHAP & SHapley Additive exPlanations\\
eFAST & Extended Fourier Amplitude Sensitivity Test\\
MAE & Mean Absolute Error\\
MSE & Mean Squared Error\\
RMSE & Root Mean Squared Error\\
MSLE & Mean Squared Logarithmic Error\\
RMSLE & Root Mean Squared Logarithmic Error\\
MLP & Multilayer-Perceptron\\
PCE & Polynomial Chaos Expansion\\
\end{tabular}
}

\appendixtitles{no} 
\appendixstart
\appendix

\begin{adjustwidth}{-\extralength}{0cm}

\bibliography{bibl}

\reftitle{References}

\end{adjustwidth}
\end{document}